%% file: main.tex
\pdfoutput=1
\documentclass[10pt,twocolumn,letterpaper]{article}
\usepackage{wacv}
\usepackage{times}
\usepackage{epsfig}
\usepackage{graphicx}
\usepackage{amsmath}
\usepackage{amssymb}

\def\wacvPaperID{231} 

\wacvfinalcopy 

\ifwacvfinal
\def\assignedStartPage{1} 
\fi
\usepackage[pagebackref=true,breaklinks=true,colorlinks,bookmarks=false]{hyperref}

\ifwacvfinal
\else
\fi

\usepackage[misc]{ifsym}
\begin{document}

\title{Compositional Learning of Image-Text Query for Image Retrieval}


\author{
Muhammad Umer Anwaar (\Letter) $^{1,2*}$, Egor Labintcev$^{1,2}$ \thanks{ indicates equal contribution. Umer contributed the rotation in complex space idea, complex projection module and rotationally symmetric loss. Egor and Umer contributed the idea of fusing different modalities before concatenation. Egor came up with the reconstruction loss idea and contributed by testing hypotheses and running numerous experiments.}, Martin Kleinsteuber$^{1,2}$\\
${^1}$Technische Universit{\"a}t M{\"u}nchen, Germany\\
${^2}$Mercateo AG, Germany\\
\{umer.anwaar, egor.labintcev\}@tum.de, martin.kleinsteuber@mercateo.com
}

\maketitle

\begin{abstract}

In this paper, we investigate the problem of retrieving images from a database
based on a multi-modal (image-text) query. Specifically, the query text prompts some modification in the query image and the task is to retrieve images with the desired modifications. For instance, a user of an E-Commerce platform is interested in buying a dress, which should look similar to her friend's dress, but the dress should be of white color with a ribbon sash. In this case, we would like the algorithm to retrieve some dresses with desired modifications in the query dress.
We propose an autoencoder based model, ComposeAE, to learn the composition of image and text query for retrieving images.
We adopt a deep metric learning approach and learn a metric that pushes composition of source image and text query closer to the target images.
We also propose a rotational symmetry constraint on the optimization problem.
Our approach is able to outperform the state-of-the-art method TIRG \cite{TIRG} on three benchmark datasets, namely: MIT-States, Fashion200k and Fashion IQ.
In order to ensure fair comparison, we introduce strong baselines by enhancing
TIRG method.
To ensure reproducibility of the results, we publish our code here: \url{https://github.com/ecom-research/ComposeAE}.
\end{abstract}

\vspace{-5mm}
\section{Introduction}

\begin{figure}
\centering
\includegraphics[width=0.7\linewidth]{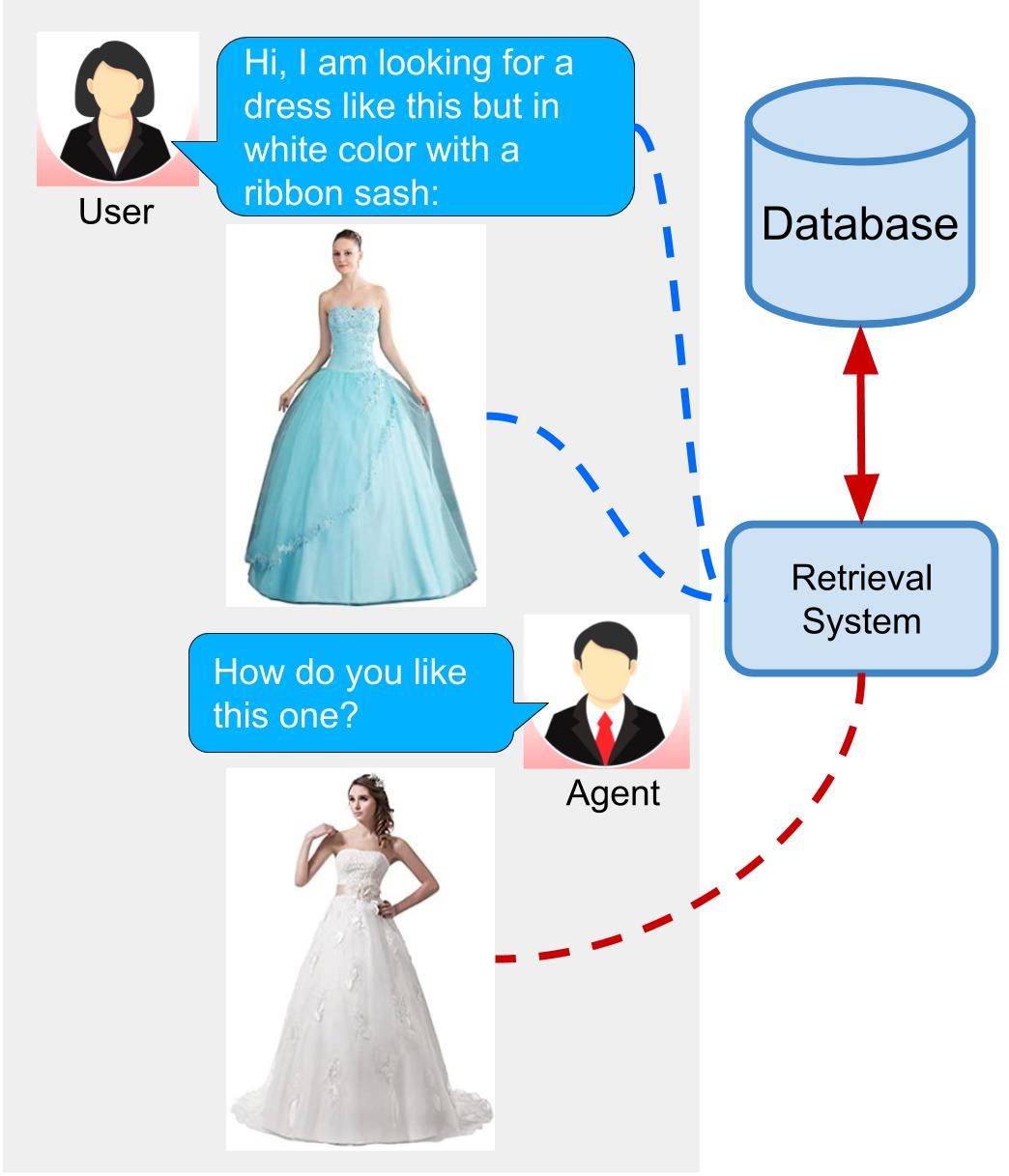}
\caption{Potential application scenario of this task }
\label{Teaser}
\end{figure}

One of the peculiar features of human perception is multi-modality. We unconsciously attach attributes to objects, which can sometimes uniquely identify them.
For instance, when a person says \textit{apple} it is quite natural that an image of an apple, which may be green or red in color, forms in their mind.
In information retrieval, the user seeks information from a retrieval system by sending a query. Traditional information retrieval systems allow a unimodal query, i.e., either a text or an image.
Advanced information retrieval systems should enable the users in expressing the concept in their mind by allowing a multi-modal query.

In this work, we consider such an advanced retrieval system, where users can retrieve images from a database based on a multi-modal query. Concretely, we have an image retrieval task where the input query is specified in the form of an image and natural language expressions describing the desired modifications in the query image. Such a retrieval system offers a natural and effective interface \cite{saha2018towards}.
This task has applications in the domain of E-Commerce search, surveillance systems and internet search.
Fig.~\ref{Teaser} shows a potential application scenario of this task.


Recently, Vo \etal \cite{TIRG} have proposed the \textit{Text Image Residual Gating} (TIRG) method for composing the query image and text for image retrieval. They have achieved state-of-the-art (SOTA) results on this task.
However, their approach does not perform well for real-world application scenarios, i.e.~with long and detailed texts (see Sec.~\ref{results}).
We think the reason is that their approach is too focused on changing the image space and does not give the query text its due importance. The gating connection takes element-wise product of query image features with image-text representation after passing it through two fully connected layers.
In short, TIRG assigns huge importance to query image features by putting it directly in the final composed representation.
Similar to \cite{santoro2017simple,vinyals2015show}, they employ LSTM for extracting features from the query text. This works fine for simple queries but fails for more realistic queries.

In this paper, we attempt to overcome these limitations by proposing ComposeAE, an autoencoder based approach for composing the modalities in the multi-modal query.
We employ a pre-trained BERT model \cite{devlin2019bert} for extracting text features, instead of LSTM. We hypothesize that by jointly conditioning on both left and right context, BERT is able to give better representation for the complex queries.
Similar to TIRG \cite{TIRG}, we use a pre-trained ResNet-17 model for extracting image features.
The extracted image and text features have different statistical properties as they are extracted from independent uni-modal models.
We argue that it will not be beneficial to fuse them by passing through a few fully connected layers, as typically done in image-text joint embeddings \cite{ wang2016learning}.

We adopt a novel approach and map these features to a complex space.
We propose that the target image representation is an element-wise rotation of the representation of the source image in this complex space.
The information about the degree of rotation is specified by the text features.
We learn the composition of these complex vectors and their mapping to the target image space by adopting a deep metric learning (DML) approach.
In this formulation, text features take a central role in defining the relationship between query image and target image.
This also implies that the search space for learning the composition features is restricted.
From a DML point of view, this restriction proves to be quite vital in learning a good similarity metric.

We also propose an explicit rotational symmetry constraint on the optimization problem based on our novel formulation of composing the image and text features.
Specifically, we require that multiplication of the target image features with the complex conjugate of the query text features should yield a representation similar to the query image features. We explore the effectiveness of this constraint in our experiments (see Sec.~\ref{ablation}).

We validate the effectiveness of our approach on three datasets: MIT-States, Fashion200k and Fashion IQ.
In Sec.~\ref{experiments}, we show empirically that ComposeAE is able to learn a better composition of image and text queries and outperforms SOTA method.
In DML, it has been recently shown that improvements in reported results are exaggerated and performance comparisons are done unfairly \cite{musgrave2020metric}. In our experiments, we took special care to ensure fair comparison. For instance, we introduce several variants of TIRG. Some of them show huge improvements over the original TIRG.
We also conduct several ablation studies to quantify the contribution of different modules in the improvement of the ComposeAE performance.

Our main contributions are summarized below:
\begin{itemize}
\item We propose a ComposeAE model to learn the composed representation of image and text query.
\item We adopt a novel approach and argue that the source image and the target image lie in a common complex space. They are rotations of each other and the degree of rotation is encoded via query text features.
\item We propose a rotational symmetry constraint on the optimization problem.
\item ComposeAE outperforms the SOTA method TIRG by a huge margin, i.e., 30.12\% on Fashion200k and 11.13\% on MIT-States on the Recall@10 metric.
\item We enhance SOTA method TIRG \cite{TIRG} to ensure fair comparison and identify its limitations.
\end{itemize}

\section{Related Work} \label{related}
Deep metric learning (DML) has become a popular technique for solving retrieval problems.
DML aims to learn a metric such that the distances between samples of the same class are smaller than the distances between the samples of different classes.
The task where DML has been employed extensively is the cross-modal retrieval, i.e. retrieving images based on text query and getting captions from the database based on the image query \cite{wang2016learning, liu2016deepfashion, Zheng2017DualPathCI, faghri2018vse++, lee2018stacked, zhang2018deep}.

In the domain of Visual Question Answering (VQA),
many methods have been proposed to fuse the text and image inputs \cite{santoro2017simple,perez2018film,noh2016image}. We review below a few closely related methods.
Relationship \cite{santoro2017simple} is a method based on relational reasoning.
Image features are extracted from CNN and text features from LSTM to create a set of relationship features. These features are then passed through a MLP and after averaging them the composed representation is obtained.
FiLM \cite{perez2018film} method tries to ``influence" the source image by applying an affine transformation
to the output of a hidden layer in the network. In order to perform complex operations, this linear transformation needs to be applied to several hidden layers.
Another prominent method is parameter hashing \cite{noh2016image} where one of the fully-connected layers in a CNN acts as the dynamic parameter layer.

In this work, we focus on the image retrieval problem based on the image and text query.
This task has been studied recently by Vo \etal \cite{TIRG}.
They propose a gated feature connection in order to keep the composed representation of query image and text in the same space as that of the target image.
They also incorporate a residual connection which learns the similarity between concatenation of image-text features and the target image features.
Another simple but effective approach is Show and Tell\cite{vinyals2015show}. They train a LSTM to predict the next word in the sequence after it has seen the image and previous words. The final state of this LSTM is considered the composed representation.
Han \etal \cite{han2017automatic} presents an interesting approach to learn spatially-aware attributes from product description and then use them to retrieve products from the database. But their text query is limited to a predefined set
of attributes.
Nagarajan \etal \cite{nagarajan2018attributes} proposed an embedding approach, ``Attribute as Operator", where text query is embedded as a transformation matrix. The image features are then transformed with this matrix to get the composed representation.

This task is also closely related with interactive image retrieval task \cite{guo2018dialog, tan2019drill} and attribute-based product retrieval task \cite{zhao2017memory, TIRG}.
These approaches have their limitations such as that the query texts are limited to a fixed set of relative attributes \cite{zhao2017memory}, require multiple rounds of natural language queries as input \cite{guo2018dialog, tan2019drill} or that query texts can be only one word i.e. an attribute \cite{han2017automatic}. In contrast, the input query text in our approach is not limited to a fixed set of attributes and does not require multiple interactions with the user. Different from our work, the focus of these methods is on modeling the interaction between user and the agent.

\section{Methodology}
\subsection{Problem Formulation}

Let $\mathcal{X} = \{x_1,x_2, \cdots , x_n\}$ denote the set of query images, $\mathcal{T} = \{t_1,t_2, \cdots , t_n\}$ denote the set of query texts and $\mathcal{Y} = \{y_1,y_2, \cdots , y_n\}$ denote the set of target images.
Let $\psi(\cdot)$ denote the pre-trained image model, which takes an image as input and returns image features in a $d$-dimensional space.
Let $\kappa(\cdot,\cdot)$ denote the similarity kernel, which we implement as a dot product between its inputs.
The task is to learn a composed representation of
the image-text query, denoted by $g(x,t;\Theta)$, by maximising

\begin{equation}
    \max_{\Theta}  \kappa( g(x,t;\Theta), \psi(y)),
    \label{optimization}
\end{equation}
where $\Theta$ denotes all the network parameters.

\subsection{Motivation for Complex Projection} 
\label{motivation}
In deep learning, researchers aim to formulate the learning problem in such a way that the solution space is restricted in a meaningful way. This helps in learning better and robust representations.
The objective function (\autoref{optimization})
maximizes the similarity between the output of the composition function of the image-text query and the target image features.
Thus, it is intuitive to model the query image, query text and target image lying in some common space.
One drawback of TIRG is that it does not emphasize the importance of text features in defining the relationship between the query image and the target image.

Based on these insights, we restrict the compositional learning of query image and text features in such a way that: (i) query and target image features lie in the same space, (ii) text features encode the transition from query image to target image in this space and (iii) transition is symmetric, i.e. \emph{some function of the text features} must encode the reverse transition from target image to query image.

\begin{figure}
\centering
\includegraphics[width=0.5\linewidth]{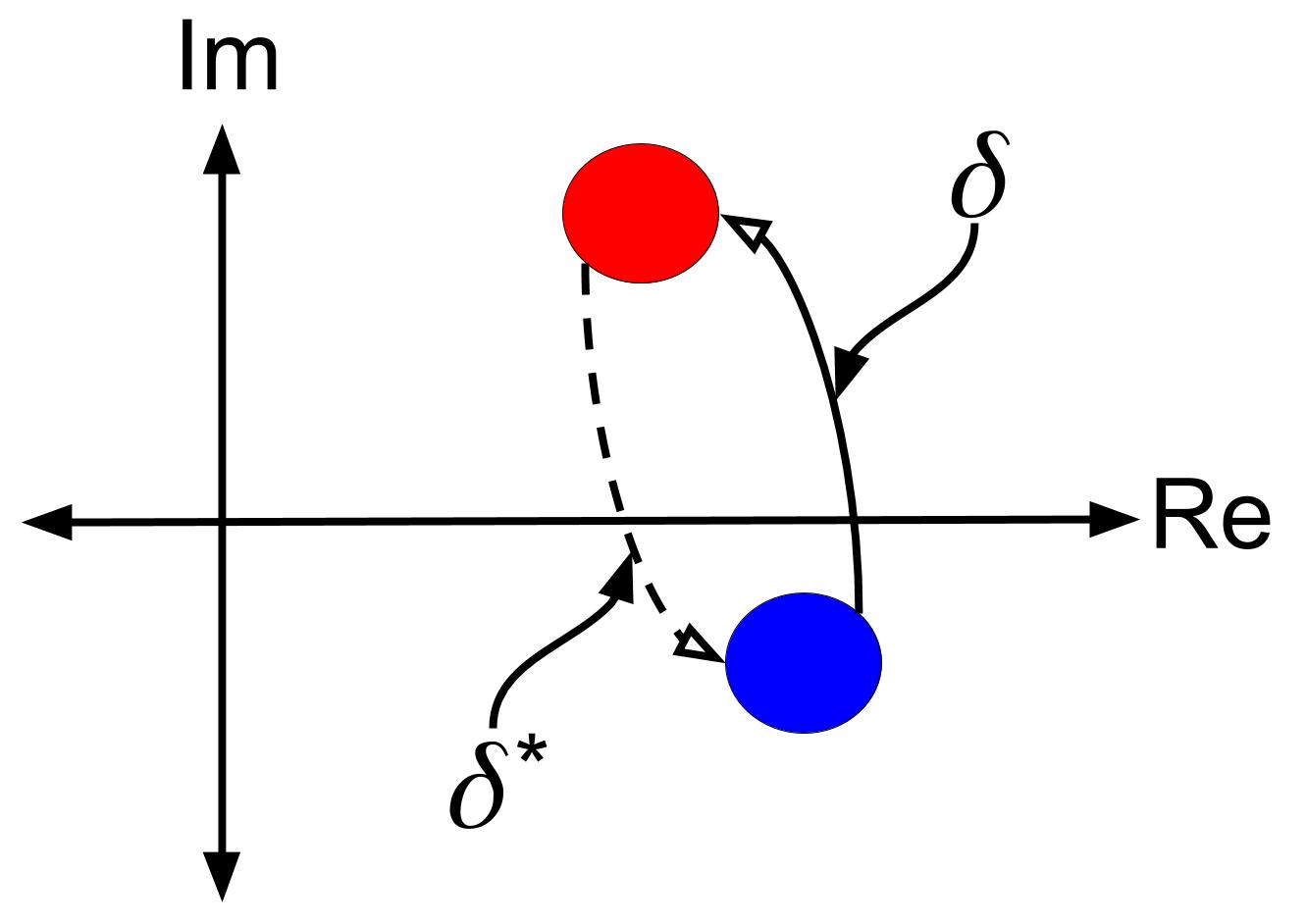}
\caption{Conceptual Diagram of Rotation of the Images in Complex Space. Blue and Red Circle represent the query and the target image respectively.
$\delta$  represents the rotation in the complex space, learned from the query text features.
$\delta^*$  represents the complex conjugate of the rotation in the complex space.}
\label{ComplexSpace}
\vspace{-3mm}
\end{figure}

\begin{figure*}
\centering
\includegraphics[width=0.759\linewidth]{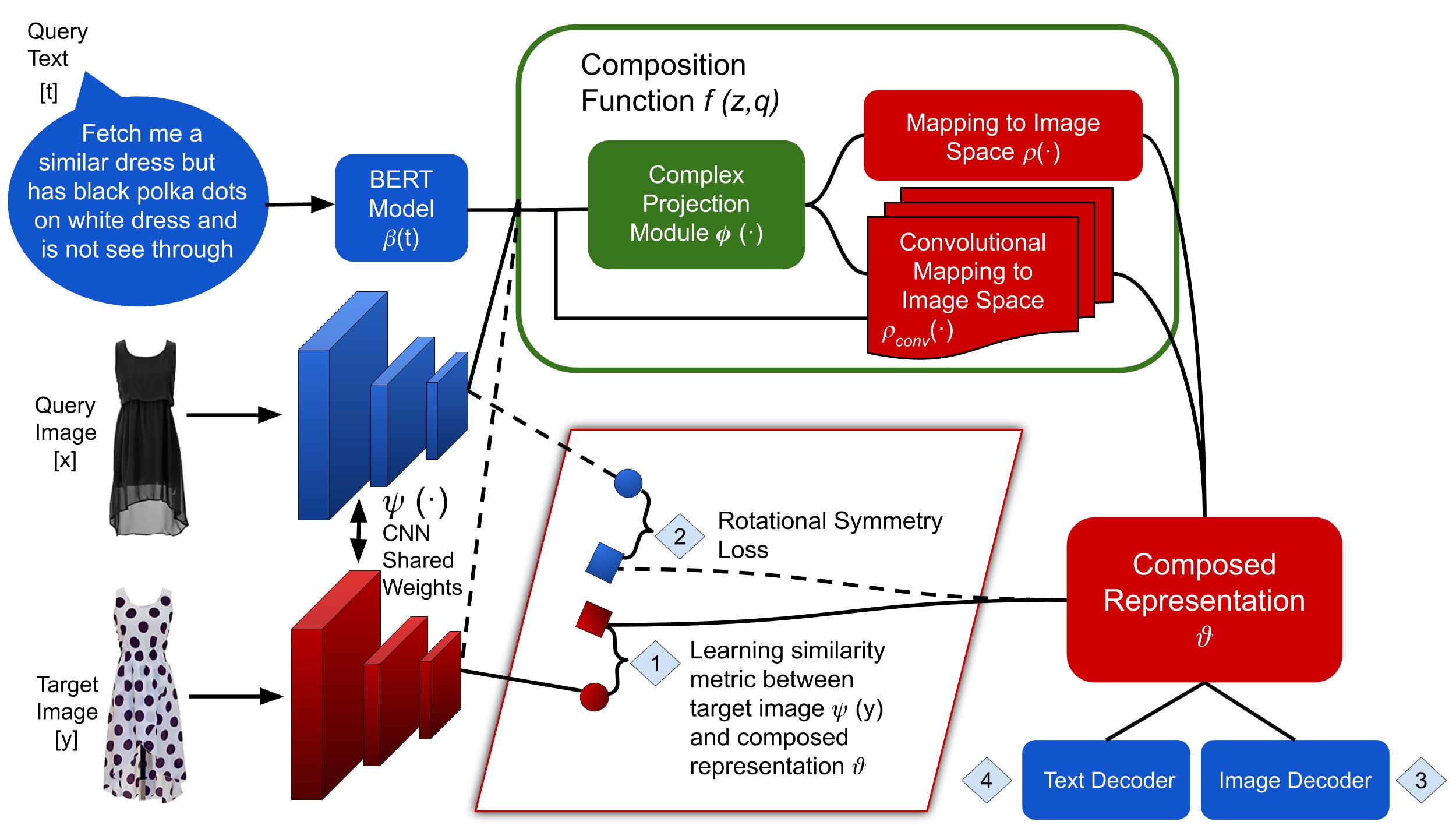}
\caption{ComposeAE Architecture: Image retrieval using text and image query. Dotted lines indicate connections needed for calculating \textit{rotational symmetry loss} (see Equations~\ref{eq: theta_conjugate}, \ref{eq:sym_loss_softmax} and \ref{eq:sym_loss_triplet}). Here 1 refers to $L_{BASE}$, 2 refers to $L^{BASE}_{SYM}$, 3 refers to $L_{RI}$ and 4 refers to $L_{RT}$.}
\label{ComposeAE_Architecture}
\end{figure*}
In order to incorporate these characteristics in the composed representation, we propose that the query image and target image are rotations (transitions) of each other in a complex space. The rotation is determined by the text features. This enables incorporating the desired text information about the image in the common complex space.
The reason for choosing the complex space is that \emph{some function of text features}
required for the transition to be symmetric can easily be defined as the complex conjugate of the text features in the complex space (see Fig.~\ref{ComplexSpace}).

Choosing such projection also enables us to define a constraint on the optimization problem, referred to as
\emph{rotational symmetry constraint} (see Equations~\ref{eq: theta_conjugate}, \ref{eq:sym_loss_softmax} and \ref{eq:sym_loss_triplet}).
We will empirically verify the effectiveness of this constraint in learning better composed representations. We will also explore the effect on performance if we fuse image and text information in the real space. Refer to Sec.~\ref{ablation}.

An advantage of modelling the reverse transition in this way is that we do not require captions of query image. This is quite useful in practice, since a user-friendly retrieval system will not ask the user to describe the query image for it.
In the public datasets, query image captions are not always available, e.g. for Fashion IQ dataset.
In addition to that, it also forces the model to learn a good ``internal" representation of the text features in the complex space.

Interestingly, such restrictions on the learning problem serve as implicit regularization. \eg, the text features only influence angles of the composed representation. This is in line with recent developments in deep learning theory \cite{Neyshabur2015InSO,neyshabur2017exploring}. Neyshabur \etal  \cite{pmlr-v40-Neyshabur15} showed that imposing simple but global constraints on the parameter space of deep networks
is an effective way of analyzing learning theoretic properties and  aids in decreasing the generalization error.


\subsection{Network Architecture}
Now we describe ComposeAE, an autoencoder based approach for composing the modalities in the multi-modal query. \autoref{ComposeAE_Architecture} presents the overview of the ComposeAE architecture.

For the image query, we extract the image feature vector living in a $d$-dimensional space, using the image model $\psi(\cdot)$ (e.g. ResNet-17), which we denote as:
\begin{equation}
    \psi(x) = z \in \mathbb{R}^d .
\end{equation}

Similarly, for the text query $t$, we extract the text feature vector living in an $h$-dimensional space, using the BERT model \cite{devlin2019bert}, $\beta(\cdot)$ as:
\begin{equation}
    \beta(t) = q \in \mathbb{R}^h .
\end{equation}

Since the image features $z$ and text features $q$ are extracted from independent uni-modal models; they have different statistical properties and follow complex distributions.
Typically in image-text joint embeddings \cite{TIRG, wang2016learning}, these features are combined using fully connected layers or gating mechanisms.

In contrast to this we propose that the source image  and target image are rotations of each other in some complex space, say, $\mathbb{C}^k$. Specifically, the target image representation is an element-wise rotation of the representation of the source image in this complex space.
The information of how much rotation is needed to get from source to target image is encoded via the query text features.
During training, we learn the appropriate mapping functions which give us the composition of $z$ and $q$ in  $\mathbb{C}^k$.

More precisely, to model the text features $q$ as specifying element-wise rotation of source image features, we learn a mapping $\gamma\colon \mathbb{R}^k \to \{D \in \mathbb{R}^{k \times k} ~|~ D \text{ is diagonal}\}$ and obtain the coordinate-wise complex rotations via
\begin{equation}
    \delta = \mathcal{E} \{j\gamma(q)\}, \nonumber
\end{equation}
where $\mathcal{E}$ denotes the matrix exponential function and $j$ is square root of $-1$.
The mapping $\gamma$ is implemented as a multilayer perceptron (MLP) i.e. two fully-connected layers with non-linear activation.

Next, we learn a mapping function, $\eta: \mathbb{R}^d \to \mathbb{C}^k$, which maps image features $z$ to the complex space. $ \eta$ is also implemented as a MLP.
The composed representation denoted by $\phi \in \mathbb{C}^k$ can be written as:
\begin{align}
    \phi = \delta \ \eta(z)
\end{align}

The optimization problem defined in Eq.~\ref{optimization} aims to maximize the similarity between the composed features and the target image features extracted from the image model.
Thus, we need to learn a mapping function, $\rho: \mathbb{C}^k \mapsto \mathbb{R}^d$, from the complex space $\mathbb{C}^k$ back to the $d$-dimensional real space where extracted target image features exist. $\rho$ is implemented as MLP.

In order to better capture the
underlying cross-modal similarity structure in the data, we learn another mapping, denoted as $\rho_{conv}$.
The convolutional mapping is implemented as two fully connected layers followed by a single convolutional layer. It learns 64 convolutional filters and adaptive max pooling is applied on them to get the representation from this convolutional mapping. This enables learning effective local interactions among different features. In addition to $\phi$, $\rho_{conv}$ also takes raw features $z$ and $q$ as input.
$\rho_{conv}$ plays a really important role for queries where the query text asks for a modification that is spatially localized. \eg, a user wants a t-shirt with a different logo on the front (see second row in Fig.~\ref{fig:fIQqual}).

Let $f(z,q)$ denote the overall composition function which learns how to effectively compose \emph{extracted image and text features} for target image retrieval.
The final representation, $\vartheta \in \mathbb{R}^d$, of the composed image-text features can be written as follows:
\begin{equation}
\vartheta = f(z,q)= a \ \rho(\phi) + b \ \rho_{conv}(\phi,z,q) ,
\end{equation}
where $a$ and $b$ are learnable parameters.

In autoencoder terminology, the encoder has learnt the composed representation of image and text query, $\vartheta$.
Next, we learn to reconstruct the extracted image $z$ and text features $q$ from $\vartheta$. Separate decoders are learned for each modality, i.e., image decoder and text decoder denoted by $d_{img}$ and $d_{txt}$ respectively.
The reason for using the decoders and reconstruction losses is two-fold: first, it acts as regularizer on the learnt composed representation and secondly, it forces the composition function to retain relevant text and image information in the final representation. Empirically, we have seen that these losses reduce the variation in the performance and aid in preventing overfitting.

\subsection{Training Objective}
We adopt a deep metric learning (DML) approach to train ComposeAE. Our training objective is to learn a similarity metric, $\kappa(\cdot,\cdot): \mathbb{R}^d \times \mathbb{R}^d \mapsto \mathbb{R}$, between composed image-text query features $\vartheta$ and extracted target image features $\psi(y)$. The composition function $f (z,q)$ should learn to map semantically similar points from the data manifold in $\mathbb{R}^d \times \mathbb{R}^h$ onto metrically close points in $\mathbb{R}^d$.
Analogously, $f(\cdot,\cdot)$ should push the composed representation away from  non-similar images in $\mathbb{R}^d$.

For sample $i$ from the training mini-batch of size $N$, let $\vartheta_i$ denote the composition feature, $\psi(y_i)$ denote the target image features and $\psi(\Tilde{y}_i)$ denote the randomly selected negative image from the mini-batch.
We follow TIRG \cite{TIRG} in choosing the base loss for the datasets.

So, for MIT-States dataset, we employ triplet loss with soft margin as a base loss. It is given by:
\begin{align}
L_{ST} =  \frac{1}{MN} \sum_{i=1}^N \sum_{m=1}^M \log\Big\{1 \!+\!  \exp\{\kappa(\vartheta_i,\! \psi(\Tilde{y}_{i,m})) \nonumber \\
\ -\ \kappa(\vartheta_i,\! \psi(y_i))\}\Big\},
\end{align}
where $M$ denotes the number of triplets for each training sample $i$. In our experiments, we choose the same value as mentioned in the TIRG code, i.e. 3.

For Fashion200k and Fashion IQ datasets, the base loss is the softmax loss with similarity kernels, denoted as $L_{SMAX}$. For each training sample $i$, we normalize the similarity between the composed query-image features ($\vartheta_i$) and target image features by dividing it with the sum of similarities between $\vartheta_i$ and all the target images in the batch. This is equivalent to the classification based loss in \cite{TIRG,goldberger2005neighbourhood, snell2017prototypical, movshovitz2017no}.

\begin{align}
\label{eq:loss_softmax}
L_{SMAX} =  \frac{1}{N}  \sum_{i=1}^N - \log \Bigg \{\frac{\exp\{\kappa(\vartheta_i, \psi(y_i))\}}
{\sum_{j=1}^N \exp\{\kappa(\vartheta_i, \psi(y_j))\}} \Bigg\},
\end{align}

In addition to the base loss, we also incorporate two reconstruction losses in our training objective. They act as regularizers on the learning of the composed representation. The image reconstruction loss is given by:

\begin{align}
\label{eq:recon_img}
L_{RI} =  \frac{1}{N}  \sum_{i=1}^N  \Big\| z_i - \hat{z_i}  \Big\|^2_2,
\end{align}
where $\hat{z_i} = d_{img}(\vartheta_i)$.

Similarly, the text reconstruction loss is given by:
\begin{align}
\label{eq:recon_txt}
L_{RT} =  \frac{1}{N}  \sum_{i=1}^N  \Big\| q_i - \hat{q_i}  \Big\|^2_2,
\end{align}
where $\hat{q_i} = d_{txt}(\vartheta_i)$.

\subsubsection{Rotational Symmetry Loss}
As discussed in \autoref{motivation}, based on our novel formulation of learning the composition function, we can include a \emph{rotational symmetry loss} in our training objective.
Specifically, we require that the composition of the target image features with the complex conjugate of the text features should be similar to the query image features.
In concrete terms, first we obtain
the complex conjugate of the text features projected in the complex space. It is given by:
\begin{equation}
    \delta^* = \mathcal{E}\{-j\gamma(q)\}.
\end{equation}

Let $\Tilde{\phi}$ denote the composition of $\delta^*$ with the target image features $\psi(y)$ in the complex space. Concretely:
\begin{align}
    \Tilde{\phi} &=  \delta^* \ \eta(\psi(y)) 
\end{align}
Finally, we compute the composed representation, denoted by $\vartheta^*$, in the following way:
\begin{align}
\vartheta^* = f (\psi(y),q)= a \ \rho(\Tilde{\phi}) + b \ \rho_{conv}(\Tilde{\phi},\psi(y),q)
\label{eq: theta_conjugate}
\end{align}
The \emph{rotational symmetry constraint} translates to maximizing this similarity kernel: $\kappa(\vartheta^*, z)$.
We incorporate this constraint in
our training objective by employing softmax loss or soft-triplet loss depending on the dataset.

Since for Fashion datasets, the base loss is $L_{SMAX}$, we calculate the rotational symmetry loss, $L_{SYM}^{SMAX}$, as follows:
\begin{align}
\label{eq:sym_loss_softmax}
L_{SYM}^{SMAX} =  \frac{1}{N}  \sum_{i=1}^N - \log \Bigg \{\frac{\exp\{\kappa(\vartheta^*_i, z_i)\}}
{\sum_{j=1}^N \exp\{\kappa(\vartheta^*_i, z_j)\}} \Bigg\},
\end{align}

Analogously, the resulting loss function, $L_{SYM}^{ST}$, for MIT-States is given by:
\begin{align}
\label{eq:sym_loss_triplet}
L_{SYM}^{ST} =  \frac{1}{MN} \sum_{i=1}^N \sum_{m=1}^M \log\Big\{1 \!+\!  \exp\{\kappa(\vartheta^*_i,\! \Tilde{z}_{i,m}) \nonumber \\
\ -\ \kappa(\vartheta^*_i,\! z_i)\}\Big\},
\end{align}

The total loss is computed by the weighted sum of above mentioned losses. It is given by:
\begin{align}
\label{total_loss}
L_{T} = L_{BASE} + \lambda_{SYM} \ L_{SYM}^{BASE} +  \lambda_{RI} \  L_{RI} +  \lambda_{RT} \  L_{RT} ,
\end{align}
where $BASE \in \{SMAX,ST\}$ depending on the dataset.

\section{Experiments}
 \label{experiments}
\subsection{Experimental Setup}

We evaluate our approach on three real-world datasets, namely:  MIT-States\cite{StatesAndTransformations},  Fashion200k \cite{han2017automatic} and Fashion IQ \cite{guo2019fashion}. For evaluation, we follow the same protocols as other recent works \cite{TIRG,han2017automatic,perez2018film}.
We use recall at rank $k$, denoted as $R@k$, as our evaluation metric.
We repeat each experiment 5 times in order to estimate the mean and the standard deviation in the performance of the models.

To ensure fair comparison, we keep the same hyperparameters as TIRG \cite{TIRG}
and use the same optimizer (SGD with momentum).
Similar to TIRG, we use ResNet-17 for image feature extraction to get 512-dimensional feature vector.
In contrast to TIRG, we use pretrained BERT \cite{devlin2019bert} for encoding text query.
Concretely, we employ BERT-as-service \cite{xiao2018bertservice} and use Uncased BERT-Base which outputs a 768-dimensional feature vector for a text query.
Further implementation details can be found in the code: \url{https://github.com/ecom-research/ComposeAE}.

\subsection{Baselines} \label{baseline}
We compare the results of ComposeAE with several methods, namely:
Show and Tell,
Parameter Hashing,
Attribute as Operator,
Relationship,
FiLM and
TIRG. We explained them briefly in Sec.~\ref{related}.

In order to identify the limitations of TIRG and to ensure fair comparison with our method, we introduce three variants of TIRG. First, we employ the BERT model as a text model instead of LSTM, which will be referred to as \textit{TIRG with BERT}.
Secondly, we keep the LSTM but text query contains full target captions.
We refer to it as \textit{TIRG with Complete Text Query}.
Thirdly, we combine these two variants and get \textit{TIRG with BERT and Complete Text Query}.
The
reason for complete text query baselines is that the
original
TIRG approach \textit{generates} text query by finding one word difference in the source and target image captions.
It disregards all other words in the target captions.

While such formulation of queries may be effective on some datasets, but the restriction on the specific form (or length) of text query largely constrain the information that a user can convey to benefit the retrieval process.
Thus, such an approach of generating text query has limited applications in real life scenarios, where a user usually describes the modification text with multiple words.
This argument is also supported by several recent studies \cite{guo2019fashion,guo2018dialog, tan2019drill}.
In our experiments, Fashion IQ dataset contains queries \textit{asked by humans} in natural language, with an average length of 13.5 words. (see Table~\ref{tab:Datasets}). Due to this reason, we can not get results of original TIRG on this dataset.

\begin{table}
\centering

\scalebox{0.9}
{\begin{tabular}{|l|c|c|c|}
\hline
& MIT-States&Fashion200k&Fashion IQ \\
\hline\hline
Total images  & 53753 & 201838 & 62145 \\
\hline
\# train queries & 43207 & 172049 & 46609 \\
\hline
\# test queries & 82732 & 33480 & 15536 \\
\hline
Average length of &&& \\
complete text query &2&4.81&13.5 \\
\hline
Average \# of  &&&\\
target images  & 26.7 & 3 & 1 \\
per query &&&\\
\hline
\end{tabular}}
\vspace{3mm}
\caption{Dataset statistics}
\label{tab:Datasets}
\end{table}
\subsection{Datasets}
Table~\ref{tab:Datasets} summarizes the statistics of the datasets.
The train-test split of the datasets is the same for all the methods.

\noindent\textbf{MIT-States} \cite{StatesAndTransformations} dataset consists of  $\sim$60k diverse real-world images where each image is described by an adjective (state) and a noun (categories), e.g. ``ripe tomato".
There are 245 nouns in the dataset and 49 of them are reserved for testing.
This split ensures that the algorithm is able to learn the composition on the unseen nouns (categories).
The input image (say ``unripe tomato") is sampled and the text query asks to change the state to ripe. The algorithm is considered successful if it retrieves the correct target image (``ripe tomato") from the pool of \textit{all test images}.
\begin{table}
\centering
\begin{tabular}{|l|ccc|}
\hline
Method & R@1 & R@5 & R@10\\
\hline\hline
Show and Tell & 11.9$^{\pm0.1}$ & 31.0$^{\pm0.5}$ & 42.0$^{\pm0.8}$ \\
Att. as Operator  & 8.8$^{\pm0.1}$ & 27.3$^{\pm0.3}$ & 39.1$^{\pm0.3}$ \\
Relationship  & 12.3$^{\pm0.5}$ & 31.9$^{\pm0.7}$ & 42.9$^{\pm0.9}$ \\
FiLM  & 10.1$^{\pm0.3}$ & 27.7$^{\pm0.7}$ & 38.3$^{\pm0.7}$ \\
TIRG & 12.2$^{\pm0.4}$ & 31.9$^{\pm0.3}$ & 43.1$^{\pm0.3}$\\
\hline

TIRG with BERT & {12.3}$^{\pm0.6}$ & {31.8}$^{\pm0.3}$ & {42.6}$^{\pm0.8}$  \\
TIRG with &  &  & \\ Complete Text Query & 7.9$^{\pm 1.9}$ & 28.7$^{\pm 2.5}$ & 34.1$^{\pm2.9}$  \\
\hline

TIRG with BERT and &&&\\ Complete Text Query & \underline{13.3}$^{\pm0.6}$ & \underline{34.5}$^{\pm1.0}$ & \underline{46.8}$^{\pm1.1}$ \\
\hline
ComposeAE & \textbf{13.9}$^{\pm0.5}$ & \textbf{35.3}$^{\pm0.8}$ & \textbf{47.9}$^{\pm0.7}$\\ 

\hline
\end{tabular}
\vspace{2mm}
\caption{Model performance comparison on MIT-States. The best number is in bold and the second best is underlined.}
\label{tab:mitstates}
\end{table}

\noindent\textbf{Fashion200k} \cite{han2017automatic} consists of $\sim$200k images of 5 different fashion categories, namely: pants, skirts, dresses, tops and jackets.
Each image has a human annotated caption, e.g. ``blue knee length skirt".

\label{fiq}
\noindent\textbf{Fashion IQ}\cite{guo2019fashion}
is a challenging dataset consisting
of 77684 images belonging to three categories: dresses, top-tees and shirts.
Fashion IQ has two human written annotations for each target image. We report the performance on the validation set as the test set labels are not available.

\subsection{Discussion of Results} \label{results}

Tables~\ref{tab:mitstates},~\ref{tab:f200k} and \ref{tab:fiq} summarize the results of the performance comparison. In the following, we discuss these results to gain some important insights into the problem.

First, we note that our proposed method ComposeAE outperforms other methods by a significant margin.
On Fashion200k, the performance improvement of ComposeAE over the original TIRG and its enhanced variants is most significant.
Specifically, in terms of R@10 metric, the performance improvement over the second best method is 6.96\% and 30.12\% over the original TIRG method .
Similarly on R@10, for MIT-States, ComposeAE outperforms the second best method by 2.35\% and by 11.13\% over the original TIRG method.
For the Fashion IQ dataset
, ComposeAE has 2.61\% and 3.82\% better performance than the second best method in terms of R@10 and R@100 respectively.

\begin{table}[ht]

\centering

{\begin{tabular}{|l|ccc|}
\hline
Method & R@1 & R@10 & R@50 \\
\hline\hline
Han \etal~\cite{han2017automatic}  & 6.3 & 19.9 & 38.3 \\
Show and Tell & 12.3$^{\pm1.1}$ & 40.2$^{\pm1.7}$ & 61.8$^{\pm0.9}$ \\
Param Hashing & 12.2$^{\pm1.1}$ & 40.0$^{\pm1.1}$ & 61.7$^{\pm0.8}$ \\
Relationship & {13.0}$^{\pm0.6}$ & {40.5}$^{\pm0.7}$ & 62.4$^{\pm0.6}$ \\
FiLM  & 12.9$^{\pm0.7}$ & 39.5$^{\pm2.1}$ & 61.9$^{\pm1.9}$ \\
TIRG & {14.1}$^{\pm0.6}$ & {42.5}$^{\pm0.7}$ & {63.8}$^{\pm0.8}$ \\
\hline
TIRG with BERT & 14.2$^{\pm1.0}$ & 41.9$^{\pm1.0}$ & 63.3$^{\pm0.9}$  \\
TIRG with &  &  & \\ Complete Text Query & 18.1$^{\pm1.9}$ & \underline{52.4}$^{\pm2.7}$ & \underline{73.1}$^{\pm2.1}$  \\
\hline
TIRG with BERT and &  &  & \\ Complete Text Query & \underline{19.9}$^{\pm1.0}$ & {51.7}$^{\pm1.5}$ & {71.8}$^{\pm1.3}$  \\
\hline
ComposeAE & \textbf{22.8}$^{\pm0.8}$ & \textbf{55.3}$^{\pm0.6}$ & \textbf{73.4}$^{\pm0.4}$ \\
\hline
\end{tabular}}
\vspace{2mm}
\caption{Model performance comparison on Fashion200k. The best number is in bold and the second best is underlined.}
\label{tab:f200k}
\end{table}
\begin{figure*}
\centering
\includegraphics[width=0.89\linewidth]{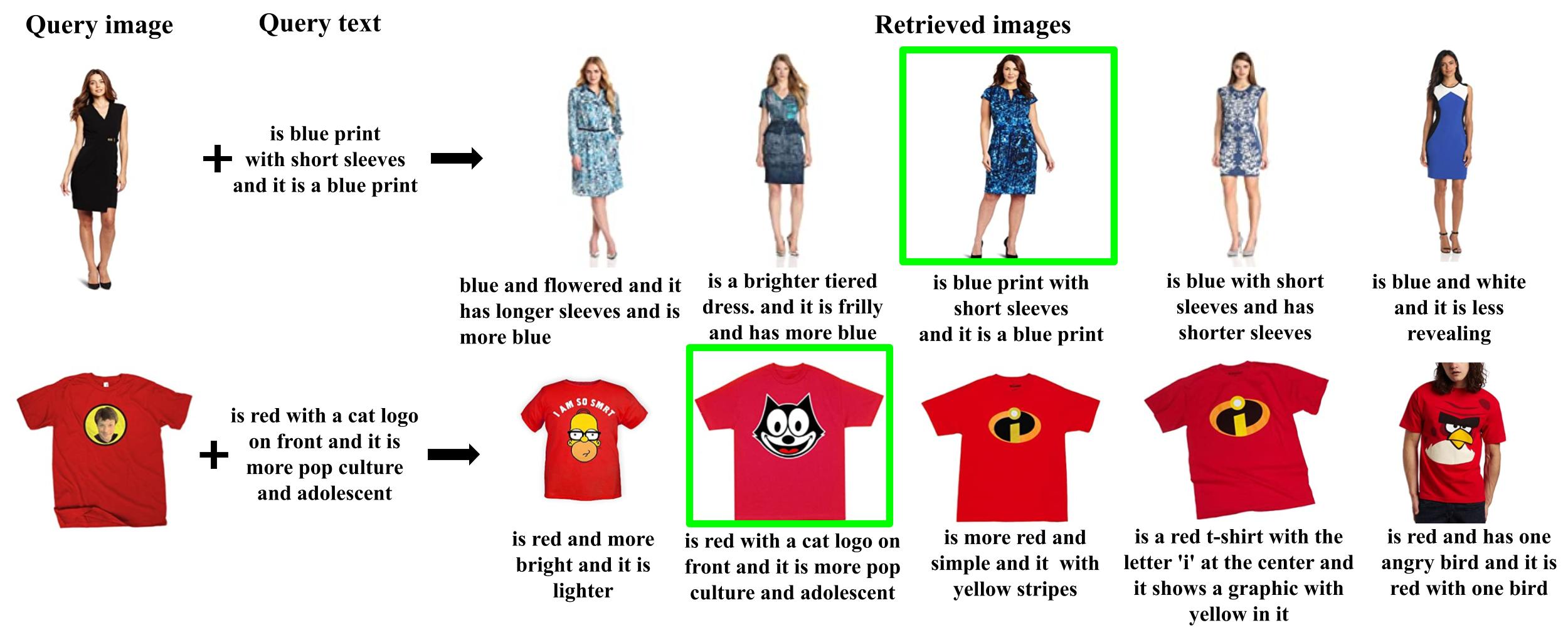}
\caption{Qualitative Results: Retrieval examples from FashionIQ Dataset}
\label{fig:fIQqual}
\end{figure*}
\begin{table}[ht]

\centering
\begin{tabular}{|l|ccc|}
\hline
Method & R@10 & R@50 & R@100 \\
\hline\hline
TIRG with &  &  & \\ Complete Text Query & 3.34$^{\pm0.6}$  & 9.18$^{\pm0.9}$  &  9.45$^{\pm0.8}$ \\
\hline
TIRG with BERT and &  &  & \\ Complete Text Query & \underline{11.5}$^{\pm0.8}$ & \underline{28.8}$^{\pm1.5}$ & \underline{28.8}$^{\pm1.6}$
\\
\hline
ComposeAE & \textbf{11.8}$^{\pm0.9}$ & \textbf{29.4}$^{\pm1.1}$ & \textbf{29.9}$^{\pm1.3}$ \\

\hline
\end{tabular}
\vspace{2mm}
\caption{Model performance comparison on Fashion IQ. The best number is in bold and the second best is underlined.}
\label{tab:fiq}
\end{table}
Second, we observe that the performance of the methods on MIT-States and Fashion200k datasets is in a similar range as compared to the range on the Fashion IQ. For instance, in terms of R@10, the performance of \textit{TIRG with BERT and Complete Text Query} is 46.8 and 51.8 on MIT-States and Fashion200k datasets while it is 11.5 for Fashion IQ.
The reasons which make
Fashion IQ the most challenging among the three datasets are:
(i) the text query is quite complex and detailed and (ii) there is only one target image per query (See Table \ref{tab:Datasets}).
That is even though the algorithm retrieves semantically similar images
but they will not be considered correct by the recall metric.
For instance, for the first query in Fig.\ref{fig:fIQqual}, we can see that the second, third and fourth image are semantically similar and modify the image as described by the query text. But if the third image which is the labelled target image did not appear in top-5, then R@5 would have been zero for this query. This issue has also been discussed in depth by Nawaz et al.\cite{nawaz2019cross}.

Third, for MIT-States and Fashion200k datasets, we observe that the TIRG variant which replaces LSTM with BERT as a text model results in slight degradation of the performance. On the other hand, the performance of the TIRG variant which uses complete text (caption) query is quite better than the original TIRG.
However, for the Fashion IQ dataset which represents a real-world application scenario, the performance of TIRG with complete text query is significantly worse.
Concretely, TIRG with complete text query performs 253\% worse than ComposeAE on R@10.
The reason for this huge variation is that the average length of complete text query for MIT-States and Fashion200k datasets is 2 and 3.5 respectively.
Whereas average length of complete text query for Fashion IQ is 12.4.
It is because TIRG uses the LSTM model and the composition is done in a way which underestimates the importance of the text query.
This shows that TIRG approach does not perform well when the query text description is more realistic and complex.

Fourth, one of the baselines (\textit{TIRG with BERT and Complete Text Query}) that we introduced shows significant improvement over the original TIRG.
Specifically, in terms of R@10, the performance gain over original TIRG is 8.58\% and 21.65\% on MIT-States and Fashion200k respectively.
This method is also the second best performing method on all datasets.
We think that with more detailed text query, BERT is able to give better representation of the query and this in turn helps in the improvement of the performance.

\noindent\textbf{Qualitative Results}:
Fig.\ref{fig:fIQqual} presents some
qualitative retrieval results for Fashion IQ. For the first query, we see that
all images are in ``blue print" as requested by text query. The second request in the text query was that the dress should be ``short sleeves", four out of top-5 images fulfill this requirement.
For the second query, we can observe that all retrieved images share the same semantics and are visually similar to the target images.
Qualitative results for other two datasets are given in the supplementary material.
\input{ablation_arch_loss}
\vspace{-3mm}
\subsection{Ablation Studies} \label{ablation}
We have conducted various ablation studies, in order to gain insight into which parts of our approach helps in the high performance of ComposeAE.
Table~\ref{tab:ablation} presents the quantitative results of these studies.

\noindent\textbf{Impact of $L_{SYM}$}: on the performance can be seen on Row 2. For Fashion200k and Fashion IQ datasets, the decrease in performance is quite significant: 7.17\% and 12.38\% respectively. While for MIT-States, the impact of incorporating $L_{SYM}$ is not that significant.
It may be because the text query is quite simple in the MIT-states case, i.e.~2 words.
This needs further investigation.

\noindent\textbf{Efficacy of Mapping to Complex Space}: ComposeAE has a complex projection module, see Fig.~\ref{ComposeAE_Architecture}. We removed this module to quantify its effect on the performance. Row 3 shows that there is a drop in performance for all three datasets. This strengthens our hypothesis that it is better to map the extracted image and text features into a common complex space than simple concatenation in real space.


\noindent\textbf{Convolutional versus Fully-Connected Mapping}: ComposeAE has two modules for mapping the features from complex space to target image space, i.e., $\rho(\cdot)$ and the second with an additional convolutional layer $\rho_{conv}(\cdot)$.
Rows 4 and 5 show that the performance is quite similar for fashion datasets. While for MIT-States, ComposeAE without $\rho_{conv}(\cdot)$ performs much better. Overall, it can be observed that for all three datasets both modules contribute in improving the performance of ComposeAE.


\section{Conclusion}
In this work, we propose ComposeAE to compose the
representation of source image rotated with the modification text in a complex space. This composed representation is mapped to the target image space and a similarity metric is learned.
Based on our novel formulation of the problem, we introduce a \emph{rotational symmetry loss} in our training objective.
Our experiments on three datasets show that ComposeAE consistently outperforms SOTA method on this task.
We enhance SOTA method TIRG \cite{TIRG} to ensure fair comparison and identify its limitations.

\section*{Acknowledgments}

This work has been supported by the Bavarian Ministry of Economic Affairs, Regional Development and Energy through the \emph{WoWNet} project  IUK-1902-003// IUK625/002.

{\small
\bibliographystyle{ieee_fullname}
\bibliography{main}
}

\end{document}

%% file: ablation_arch_loss.tex
\begin{table}[ht]
\centering
\scalebox{0.85}{
\begin{tabular}{|l|ccc|}
\hline
Method     & Fashion200k & MIT-States & Fashion IQ \\
\hline\hline
ComposeAE  &   55.3    &       47.9    & 11.8    \\
 \hline
  - without $L_{SYM}$  &  51.6     &  47.6    & 10.5 \\
 - Concat in real space   &   48.4      & 46.2   & 09.8  \\ 
 \hline
 - without $\rho_{conv}$  &  52.8      &  47.1    & 10.7  \\
 - without $\rho$     &  52.2    &  45.2   &  11.1 \\
 \hline

\end{tabular}}
\vspace{2mm}
\caption{\label{tab:ablation}Retrieval performance (R@10) of ablation studies.}
\end{table}